\documentclass[11pt,a4paper]{article}
\usepackage[hyperref]{naaclhlt2019}
\usepackage{times}
\usepackage{latexsym}
\usepackage{booktabs}
\usepackage{microtype}
\usepackage{url}
\usepackage{boldline}
\usepackage{tabularx}
\usepackage{multirow}
\usepackage{url}
\usepackage{xcolor}
\usepackage{scalefnt}
\usepackage{booktabs}
\usepackage{listings}
\usepackage{sistyle}
\SIthousandsep{,}
\usepackage{graphicx}
\usepackage{tabularx}
\newcommand{\rt}[1]{\rotatebox{90}{#1}}
\usepackage{paralist}
\usepackage{amsfonts,amsmath,amssymb,amsthm}
\usepackage{xspace}
\newcommand{\eg}{\textit{e.\,g.}\xspace}

\newcommand{\ia}{\textit{i.\,a.}\xspace}

\usepackage{multirow}
\newcommand{\alacarte}{\`{A} La Carte}
\newcommand{\fasttext}{fastText}
\newcommand{\glove}{GloVe}
\newcommand{\wordtovec}{Word2vec}
\newcommand{\bagofsubstrings}{Bag-of-Substrings}
\usepackage{amsmath,amsfonts,amssymb,amsthm}
\usepackage{mathtools}
\usepackage{commath}

\aclfinalcopy

\def\aclpaperid{}

\title{Exploring Fine-Tuned Embeddings that\\ Model Intensifiers for Emotion Analysis}

\author{
Laura Bostan \and Roman Klinger\\
Institut f\"ur Maschinelle Sprachverarbeitung\\
University of Stuttgart\\
Pfaffenwaldring 5b, 70569 Stuttgart, Germany\\
\texttt{\{laura.bostan,roman.klinger\}@ims.uni-stuttgart.de}
}

\date{}

\begin{document}
\maketitle

\begin{abstract}
  Adjective phrases like ``a little bit surprised'', ``completely
  shocked'', or ``not stunned at all'' are not handled properly by
  currently published state-of-the-art emotion classification and
  intensity prediction systems which use predominantly
  non-contextualized word embeddings as input. Based on this finding,
  we analyze differences between embeddings used by these systems in
  regard to their capability of handling such cases. Furthermore, we
  argue that intensifiers in context of emotion words need special
  treatment, as is established for sentiment polarity classification,
  but not for more fine-grained emotion prediction. To resolve this
  issue, we analyze different aspects of a post-processing pipeline
  which enriches the word representations of such phrases. This
  includes expansion of semantic spaces at the phrase level and
  sub-word level followed by retrofitting to emotion lexica. We
  evaluate the impact of these steps with \alacarte{} and
  \bagofsubstrings{} extensions based on pretrained \glove,
  \wordtovec, and \fasttext{} embeddings against a crowd-sourced
  corpus of intensity annotations for tweets containing our focus
  phrases. We show that the \fasttext-based models do not gain from
  handling these specific phrases under inspection. For \wordtovec{}
  embeddings, we show that our post-processing pipeline improves the
  results by up to 8\% on a novel dataset densely populated with
  intensifiers.
\end{abstract}

\section{Introduction}
Emotion detection in text includes tasks of mapping words, sentences,
and documents to a discrete set of emotions following a psychological
model such as those proposed by \newcite{Ekman1992} and
\newcite{Plutchik1980}, or to intensity scores or continuous values of
\textit{valence--arousal--dominance} \cite{Posner2005}. The shared
task on intensity prediction for discrete classes proposed to combine
both \cite{SemEval2018Task1,wassa2017}. In this task a tweet and an
emotion are given and the goal is to determine an intensity score
between 0 and 1.

Especially, but not only in social media, users use degree adverbs
\cite[also called intensifiers][]{quirk1985contemporary}, for instance
in ``I am \textit{kinda happy}'' \textit{vs.} ``I am \textit{very
  happy}.''  to express different levels of emotion intensity. This is
a relevant task: 10\% of tweets containing an emotion word are
modified with such an adverb in the corpus we describe in
Section~\ref{sec:data}. In this paper, we challenge the assumption
that models developed for intensity prediction perform well on tweets
containing such phrases and analyze which of the established embedding
methods \wordtovec{} \cite{mikolov2013distributed}, \glove{}
\cite{pennington2014glove}, and \mbox{\fasttext} embeddings
\cite{bojanowski2016enriching} performs well when predicting
intensities for tweets containing such phrases. We will see that the
performance of the popular and fast-to-train \wordtovec{} method can
be increased with a simple postprocessing pipeline which we present in
this paper.

As a motivating example, the DeepMoji model \cite{Felbo2017} predicts
\textit{anger} for both the example sentences ``I am not angry.'' and
``I am angry.''\footnote{\url{https://deepmoji.mit.edu}}.  Using the
model by \newcite{wu2018thungn} (one of the state-of-the-art intensity
prediction models from \newcite{SemEval2018Task1}, building their
model on top of \wordtovec{} embeddings) we also obtain \textit{anger}
as having the highest intensity for both examples.  We argue that the
models should be more sensitive to the difference between
\textit{negations}, \textit{downtoners} and \textit{amplifiers}.

With this paper, we contribute to alleviate this situation in three
aspects.  Firstly, we provide an analysis of the distribution of
degree adverbs (including negations) with emotion words and show that
not all such combinations are equally common. Secondly, we perform a
crowdsourcing experiment in which we collect scores for different
combinations of degree adverbs and emotion adjectives. We use these
data, which we make publicly available, as an additional challenging
test set for the task of intensity prediction for English. Thirdly, we
use a state-of-the-art intensity prediction model \cite{wu2018thungn}
on this test set and evaluate two methods to improve these
predictions, namely the inclusion \cite{zhao2018generalizing} and
$n$-gram embeddings via \alacarte{} of additional subword information
with \bagofsubstrings{} \cite{khodak2018alacarte}. We evaluate based
on \wordtovec, \glove{} and \mbox{\fasttext} embeddings and show that
particularly the first two benefit from these changes, but to
different extents.

\section{Related Work}\label{sec:rel}
\subsection{Degree Adverbs in Linguistics}\label{sec:ling}

Adverbs that express intensity are named \textit{degree adverbs},
\textit{degree modifiers} or \textit{intensifiers}.\footnote{In this
  paper, we will use these terms interchangeably.} The entities they
intensify are located on an abstract scale of intensity
\cite{quirk1985contemporary}. The intensifiers that scale upward are
named amplifiers and are further categorised as maximizers, such as
``completely'' and ``totally'' or boosters, such as ``really'' or
``truly''. Those that scale downward are called downtoners and are
further classified as approximators, such as ``almost'' or ``kind
of'', compromisers, such as ``fairly'', ``pretty'' and ``quite'',
diminishers, such as ``slightly'' and ``a bit'', and minimizers
\cite[\ia]{quirk1985contemporary,paradis1997degree,nevalainen2002fairly}.
Further distinction of degree modifiers is concerned with the fact
that there are intensifiers that imply boundaries, such as
``totally'', ``fully'', and ``completely'' and those that do not, such
as ``very'', ``utterly'', ``pretty''
\cite{paradis1997degree,paradis2001adjectives,paradis2000s}. Finally,
in the context of discourse, there is the property of expressing
focus, which is present in the so-called \textit{focus modifiers},
such as ``only'' and ``just'', which are also further classified in
additives, such as ``also'' and ``too'' and restrictives, such as
``only'' and ``merely''
\cite{quirk1985contemporary,athanasiadou2007subjectivity}.

English degree modifiers have also long history of research in English
studies and more generally in Language Studies. Most English studies
focus on the incidence and distribution of these adverbs in different
corpora, e.g.  \newcite{peters1994degree} study letters from Early
Modern English and shows the how the distributions of boosters change
across time.  \newcite{nevalainen2008social} study the social
variation in intensifier use, with a focus on the suffix
\textit{-ly}. More recently, \newcite{napoli2017new} collect a volume
of papers that explore the process of intensification following a
corpus-based, cross-linguistic and contrastive approach. The volume
contains various works on the variation in the distribution and
incidence of the intensifiers based on sociolinguistic features and in
a diachronic fashion.  The work brings in attention intensification in
ancient languages as well as modern languages.

A more recent work investigates the differences in the use of
intensifiers and considers English speech of adults and teenagers as
corpus. It explores two maximizers in-depth, namely ``absolutely'' and
``totally'' and shows that those prove to be more ``flexible`` in the
language used by teenagers \cite{pertejo2014s}.

\subsection{Modifiers in the context of Sentiment and Emotion Analysis}

In the context of sentiment analysis the discussion of intensifiers
and negations has gained quite some attention, since those
are primarily markers of subjectivity \cite{athanasiadou2007subjectivity}.

Negations, and in particular negation cue detection (with the goal of
scope recognition) have been the research interest of
\newcite{councill2010} and \newcite{reitan2015}, who use a lexicon for
negation cue detection and a linear-chain conditional random field for
scope recognition. In the area of distributional semantics, the
investigation of word vectors with a focus on negated adjectives
\cite{ainadistributional} is complementary to our work with regards to
negation in terms of the methods and data used.  Following this
approach, one could build a distributional semantic model whose
vocabulary includes the modified phrases. In practice, each occurrence
of a modified adjective by a degree adverb could be treated as a
single token (\eg ``not happy'' would be represented as
``not{\_}happy'').  For a general overview of modality and negation in
computational linguistics we refer the interested reader to the work
by \newcite{morante2012}.

Furthermore, \newcite{zhu2014} study the effect of
negation words on sentiment and evaluate a neural composition model.
\newcite{mohammadsvet2016} create a sentiment lexicon of phrases that
include modifiers such as negators, modals, and degree adverbs. The
phrases and their constituent words are annotated manually with the
same annotation procedure we will discuss in detail. We follow this
work closely and apply the same procedures in the context of emotion
analysis.

\newcite{dragut2014} study the effect of intensifiers on the sentiment
ratings and shows that the degree adverbs do not carry an inherent
sentiment polarity but alter the degree of the polarity of the
constituents they modify.

We argue that there is not enough work on transferring the methods used
in sentiment analysis to the more fine-grained analysis of emotions,
except for \newcite{strohm2018empirical}, who limit themselves to
analysis and do not apply state-of-the-art prediction models for
handling degree adverbs, and \newcite{carrillo2013emotion} who
consider modified emotions but predict sentiment.

\section{Methods}
In the following, we explain how we create the data sets for our
analysis (Section~\ref{sec:data}) and then how we set up the
experiments to measure the impact of \alacarte{} and
\bagofsubstrings{} on the modified phrases
(Section~\ref{sec:adaptations}).

\subsection{Data Collection and Annotation}
\label{sec:data}
As a basis of our work, we create a compositional emotion lexicon for
English Twitter and retrieve crowdsourced ratings using
\textit{Best-Worst Scaling}
\cite{louviere1991best,kiritchenko2017capturing}. We show later that
these ratings are by and large independent of context and can
therefore be interpreted as a labeled emotion lexicon of compositional
phrases.\footnote{Our data is available at
  \url{https://www.ims.uni-stuttgart.de/data/modifieremotion}.}

\subsubsection{Data Collection}\label{sec:data_collection}
Each query we use to retrieve tweets consists of a pair of an
adjective with one or a combination of several degree adverbs
(intensifiers (including amplifiers and downtoners) and negations),
for instance ``not at all surprised'' or ``not very happy''. We first
generate a comprehensive list by mapping each of Ekman's fundamental
emotions \cite{Ekman1992} to their corresponding adjective
\textit{sad, happy, disgusted, afraid, surprised, angry} and augment
this list to $333$ emotion adjectives and their synonyms from the
Oxford Dictionary of English \cite{ehrlich1980oxford}, the New Oxford
American Dictionary \cite{stevenson2010new} and Macmillan Online
English
Dictionary\footnote{\url{http://www.macmillandictionaries.com/dictionary-online/}}
and further filter this list to 43 entries which are intersubjectively
agreeable. This filter step is performed via crowdsourcing on
Prolific\footnote{\url{https://prolific.ac}}, in which we asked native
speakers of English which emotion is the closest to each synonym. We
only keep those synonyms where all annotators agreed. The
inter-annotator agreement is $\kappa=0.63$ (Fleiss' $\kappa$ over 9
annotators).

The list of degree modifiers is a combination of
\newcite{quirk1985contemporary,paradis1997degree,strohm2018empirical}. From
the cartesian product of degree modifiers with emotion adjectives, we
keep those which we find at least 10 times in the general Twitter
corpus we discuss below.  That leads to $266$ phrases.

We base our analysis on a set of 32 million tweets obtained from
Twitter with the official API between March 2006 and October 2018,
using a combination of diverse search terms corresponding to isolated
emotion word synonyms, those in combination with degree adverbs, and
frequent hashtags. We filter out retweets and full quotes, tweets with
more than 140 characters and those with less than 10 tokens, as well
as those consisting of more than 30\% hashtags, links, or usernames,
which we replace by generic respective tokens otherwise. Tweets with
more than 30\% of non-ASCII characters are also removed.

\subsubsection{Annotation Procedure} For each tweet ($t$) and emotion
($e$) we obtain emotion intensity scores $s_{t,e}\in[-1,1]$ via
\textit{Best-Worst Scaling} \cite[BWS,][]{louviere1991best}. In
general with BWS, the annotators are shown a subset of a number of
items from a list and are asked to select the \textit{best and worst}
items (or most and least some given property of interest). Within our
study, we show four items at once to the annotators.  In a first
setting, we show them four tweets that contain the queries we want to
have scores assigned for. In a second setting, we show them only the
queries without the context (the tweet) in which they were found. In
both scenarios, the annotators need to select the tweet or the query
with the highest and lowest intensity of each emotion.

\begin{figure}
  \centering
  \includegraphics[width=0.5\textwidth]{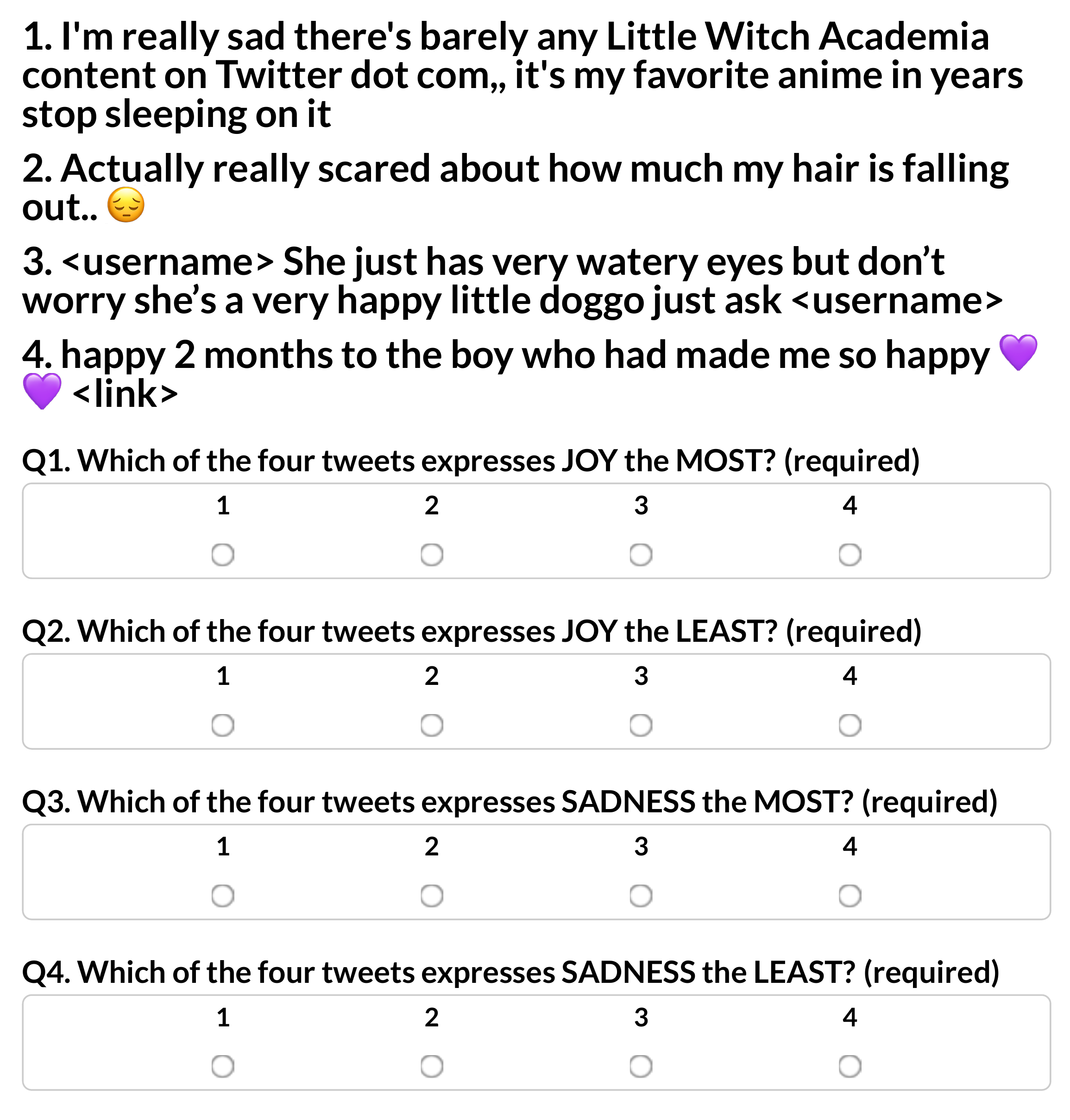}
    \caption{An example of what contributors see on the Figure Eight Platform. The 4 sentences shown are an example of a group of four tweets the contributors have to annotate. The questions Q1 to Q4 that follow below are a subset of the questionnaire. }
  \label{fig:fig8}
\end{figure}

These groups of tweets are sampled under following constraints that
have been empirically proven to lead to reliable scores
\cite{kiritchenko2018sentiment}, resulting in $532$ samples (twice the
amount of queries): (1) no two samples have the same four queries (in
any order), (2) no two queries within a sample are identical, (3) each
query occurs in 8 ($\pm 1$) different samples, (4) each pair of
queries appears in the same number of samples.
We perform two annotation experiments on the crowdsourcing platform
Figure Eight\footnote{\url{https://www.figure-eight.com}}: In
Experiment 1, we present the whole tweet to the annotator, in
Experiment 2, we only show the query phrase. This enables us to
evaluate the importance of context, shown in
Section~\ref{sec:resanno}.  Each sample was annotated by three
contributors that confirmed to be English native speakers.

\subsection{Adaptations of Embeddings}
\label{sec:adaptations}
In the following, we discuss the three methods to improve the
embeddings and later to test if these improvements add additional
information with respect to intensifiers for emotion analysis. The
evaluation will be on the downstream task of emotion intensity
prediction.

We focus on subword-level information and phrase-level information, as
those, presumably, capture intensity information.

\subsubsection{\alacarte}
With this method we learn a representation of yet unseen phrases within an
embedding space through a linear transformation of the average of the word
embeddings in the feature's contexts. The method constructs a representation
for a new phrase given a set of contexts where this phrase occurs in.

Given our Twitter corpus $\mathcal{C}_w$ consisting of contexts of words $w$
and the pre-trained word embeddings ${\bf v}_w\in\mathbb{R}^d$, of dimension
$d$, our goal is to construct a representation ${\bf v}_{q}\in\mathbb{R}^d$ of
a query $q$ given a set $\mathcal{C}_q$ of contexts it occurs in.

We learn the transform ${\bf A\in\mathbb{R}^{d\times d}}$ that can recover {\em
existing} word vectors ${\bf v}_{w}$ via \textit{linear regression} by summing
their context embeddings

\begin{equation}
  {\bf v}_{w}\approx{\bf A}\left(\frac{1}{|\mathcal{C}_{w}|}\sum\limits_{c\in\mathcal{C}_{w}}\sum\limits_{w'\in c}{\bf v}_w'\right)\,.
\end{equation} 
Using the learned transformation matrix ${\bf A}$ we can embed any new
query ${\bf v}_{q}$ in the same semantic space as the pre-trained word
embeddings via
\begin{equation}
  {\bf v}_{q}={\bf A}\left(\frac{1}{|\mathcal{C}_{q}|}\sum\limits_{c\in\mathcal{C}_{q}}\sum\limits_{w\in c}{\bf v}_w\right)\,.
\end{equation}

\subsubsection{\bagofsubstrings}
BoS generalizes pre-trained semantic spaces to unseen words. The
established approach to represent word phrases or sentences is to take
a bag of words of word embeddings.

BoS achieves its goal by first learning a mapping between the subwords
present in each word and its corresponding pre-trained vector. Then,
by using this learned subword transformation, the model is able to
generate new representations for any new word as a set of its
character $n$-grams. For us this is relevant, since we can consider
our focus phrases to be character $n$-grams instead of word $n$-grams.

Formally, the representation for a word ${\bf v}_{w}$ from the lookup
table V (which stores the embeddings of dimension $d$ for each
possible substring of length within a range) is:
\begin{equation}\label{equation:bos}
    {\bf v}_{w}  = \frac{1}{|{\mathcal{S}_{w}}|}\sum\limits_{t\in\mathcal{S}_{w}} {\bf v}_{t},
\end{equation}
where $\mathcal{S}_{w}$ is the set of each possible character
$n$-grams of length within a given range over ${w}$ and ${\bf v}_{t}$
is the vector in V indexed by t.

The model views the vector of a phrase as the average vector of all its
substrings, which are trained by minimizing the overall mean squared loss
between the generated and given vectors for each word:

\begin{equation}\label{eqn:train}
  \min_{V} \frac{1}{|W|} \sum\limits_{w\in W} l\left(\frac{1}{|{\mathcal{S}_{w}}|}\sum\limits_{t\in\mathcal{S}_{w}} {\bf v}_{t}, {\bf u}_{w}\right)
\end{equation} 
where ${\bf u}_{w}$ $\in \mathbb{R} ^{d\times |W|}$ are the target
vectors of the dimension d over the vocabulary $W$ and $l(\bf v, \bf
u) = \frac{1}{2}\norm{\bf {v} - \bf{u}}_2^2$ .

After training, similarly to the previous method, one can use the learned space
to generate a new word vector ${\bf v}_{q}$ as the average of the vectors of
all of its substrings through Equation~\ref{equation:bos}.

Since BoS produces vectors for unknown words from vectors of
substrings of characters contained in it, this allows to build vectors
for misspelled words and concatenation of words. Particularly on
Twitter data, we benefit from getting a representation for phrases
like ``sooooexcited:)'', ``verrry cheerful'', ``soo
unhappy:(''. Relevant for our analysis is that BoS uses special
characters to mark the start and the end of the word and thus helps
the model to distinguish morphemes that occur at different word parts,
like prefixes or suffixes. Through that we learn to distinguish
morphemes like ``un-'', ``-er'' and ``-est'' that are part of our
focus phrases.

Note that this method uses the same idea as in
\fasttext{}~\cite{bojanowski2016enriching}, but is for our case computationally
more efficient, since the BoS model is trained directly on top of
pre-trained vectors, instead of predicting over text corpora.

\subsubsection{Retrofitting}
We use the method of retrofitting existing embeddings
\cite{faruqui2014retrofitting} in order to enrich word vectors using
synonymity constraints provided by semantic lexicons. The algorithm
learns the word embedding matrix $A = \{{\bf {v}}_1 , {\bf {v}}_2 ,
\dots ,{\bf {v}}_n\}$ with the objective function: \begin{equation}
  \Psi (A) = \sum_{i\in V} \ [\alpha_{i} || {\bf {v}}_i -
  \hat{\bf{v}}_i ||^2 + \sum_{(i,j) \in E} \beta_{ij} ||\hat{\bf
    v}_{i} - \hat{ \bf v}_{j}||^2]
\end{equation} where an original word vector is ${\bf v}_{i}$, its synonym vector is
${\bf v}_{j}$, and inferred word vector is $\hat{\bf v}_{i}$.

Our lexicon of synonymity constraints was automatically constructed
from the data we collected in Section~\ref{sec:data_collection} by
adding an entry for each emotion adjective with its synonyms
crowdsourced as previously described.  We also added entries for the
phrases in the lexicon for retrofitting, as follows, for each emotion
phrase according to the intensifiers classification described in
Section~\ref{sec:ling}. For instance, ``not happy'' had as an entry in
the lexicon the phrases ``unhappy'' and ``not happy at all'' while
``completely cheerful'' had in its entry phrases like ``totally
cheerful'', ``totally happy'', ``completely happy'', among others
(since ``completely'' and ``totally'' are in the same class). We apply
retrofitting on phrases which origin from the extension of the space
with \alacarte.

\begin{table}
\centering
\small
\setlength\tabcolsep{0.90mm}
\begin{tabular}{lrrrrrr}
\toprule
Focus phrase & \rt{joy} & \rt{sadness} & \rt{anger} & \rt{fear} & \rt{surprise} & \rt{disgust}\\
\midrule
so happy & $+.73$  & $-.43$  & $-.50$ & $-.51$ & $-.10$ & $-.66$\\
not happy  & $-.52$ & $+.41$ & $+.02$ & $-.16$ &  $-.11$ & $+.17$ \\
kinda happy &  $+.53$ & $-.70$&  $-.67$  & $-.55$ &  $-.47$ & $-.76$ \\ 
\midrule
so sad & $-.50$ & $+.66$ & $+.04$ &  $+.13$ & $-.16$  & $+.03$ \\
not sad & $+.55$ & $-.60$ & $-.57$ & $-.55$ & $-.45$ & $-.52$ \\
kinda sad & $-.41$  & $+.62$ & $-.02$ & $+.02$ & $-.18$ & $+.02$\\
\midrule
so angry &  $-.39$ & $+.26$ & $+.86$ &  $+.21$ & $+.02$ & $+.63$\\
not angry & $+.40$ & $-.36$ & $-.82$ & $-.17$ & $-.27$ & $-.45$ \\
kinda angry &  $-.80$ & $+.68$ & $+.84$ & $+.32$ & $+.08$ & $+.64$\\
\midrule
so scared & $-.07$ & $+.15$ & $-.21$ & $+.83$ & $+.15$  & $-.13$  \\
not scared & $+.35$ & $-.35$ & $-.28$ & $-.66$ & $-.53$ & $-.33$\\
kinda scared  & $+.03$ & $+.10$  & $-.03$ & $+.71$ &  $-.13$ & $-.13$ \\
\midrule
so surprised & $+.34$ & $-.27$ & $-.09$ & $+.02$ & $+.81$ & $.00$\\
not surprised & $+.60$ & $-.56$ &  $-.50$ & $-.60$ & $-.83$  &$-.60$  \\
kinda surprised & $+.37$ & $-.37$ & $-.17$ &  $+.01$ & $+.72$ & $-.20$ \\
\midrule
so disgusted & $-.11$ & $-.02$ & $+.30$ & $+.16$ & $+.33$ & $+.88$ \\
not disgusted & $+.42$ & $-.39$ & $-.42$ & $-.36$ & $-.36$ & $-.84$\\
kinda disgusted & $-.16$ & $+.08$ & $+.41$ & $-.01$ & $+.08$  & $+.80$ \\
\bottomrule
\end{tabular}
\caption{Example queries with their BWS crowdsourced scores for the modifiers
    ``so'', ``kinda'' and the negation ``not''. For every focus phrase we have
    an intensity score between $-1$ and $+1$ for each emotion. The focus
    phrases are shown in groups made around the emotion adjectives. }
\label{tab:queries-examples} \end{table}

\section{Results} \label{sec:results}
In the following, we explore the Twitter corpus described previously, the
results of the BWS annotation of the pairs of degree adverbs and adjectives,
and finally we discuss our experimental setting and evaluation on the
downstream task of emotion intensity prediction on the different embedding
adaptation methods.

\subsection{Corpus Analysis}
\label{sec:rescorpus}
The Twitter corpus described in Section~\ref{sec:data}
contains \num{34297941} tweets out of which \num{2948397} contain emotion
phrases. Most dominant are amplifiers (49\%) followed by downtoners
($24\%$) and negations ($19\%$). Only $8\%$ contain the
emotion adjectives in superlative or comparative.

\begin{figure*}
  \centering
  \includegraphics[width=1\textwidth]{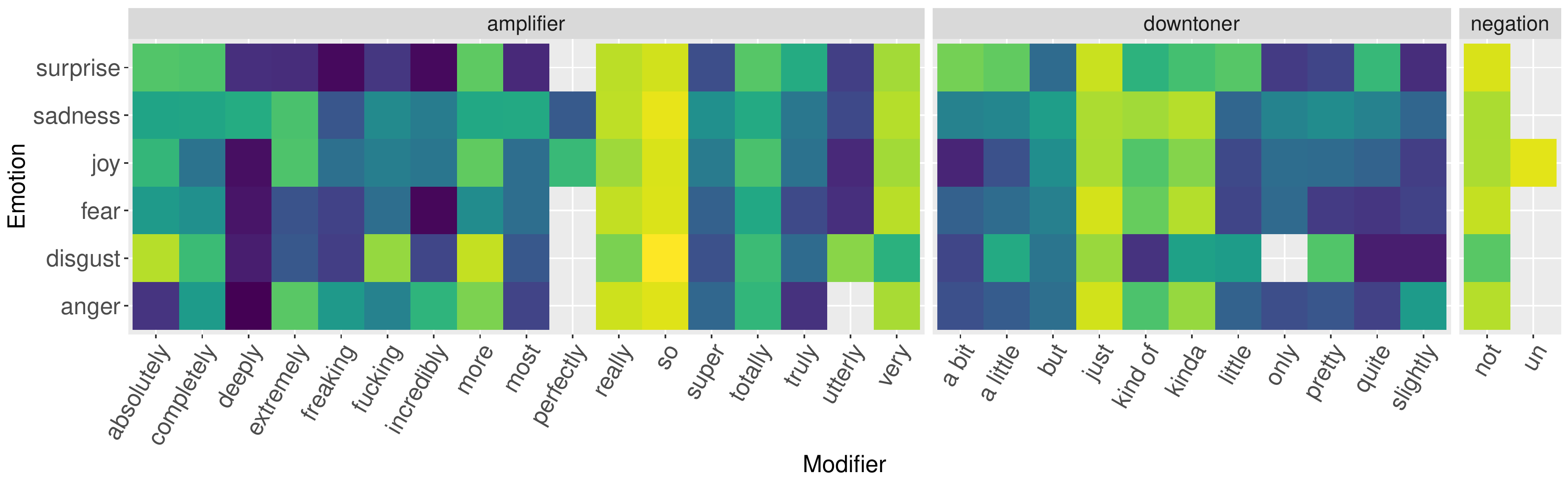}
  \caption{Relative frequencies of the most common 30 modifiers in the
    Twitter Corpus (from dark (infrequent) to yellow
    (frequent).}
  \label{fig:freq}
\end{figure*}

Figure~\ref{fig:freq} shows how often the top 30 modifiers are used
with adjectives from the basic set of emotions. We see that
\textit{disgust} is rarely downtoned and \textit{anger},
\textit{sadness}, and \textit{surprise} are amplified most
often. \textit{Joy} and \textit{fear} are relatively equally
amplified, with \textit{joy} being more negated and \textit{fear}
being more downtoned. The amplifiers ``so'' and ``really'', as well as
the downtoners ``just'' and ``kind of/kinda'' are frequently used. The
downtoner ``just'' is the most frequently used downtoner and acts at
times as an amplifier, which could explain its frequent use. We
hypothesize that this is due to their use as fillers and their
grammaticalization \cite[cf.][]{tagliamonte2006so}. Most frequently
downtoned emotion is \textit{surprise} (which is often used in phrases
like ``a little surprised'', ``quite surprised'', ``a bit
surprised'').

\begin{figure}[t]
   \centering
   \includegraphics[width=0.9\linewidth]{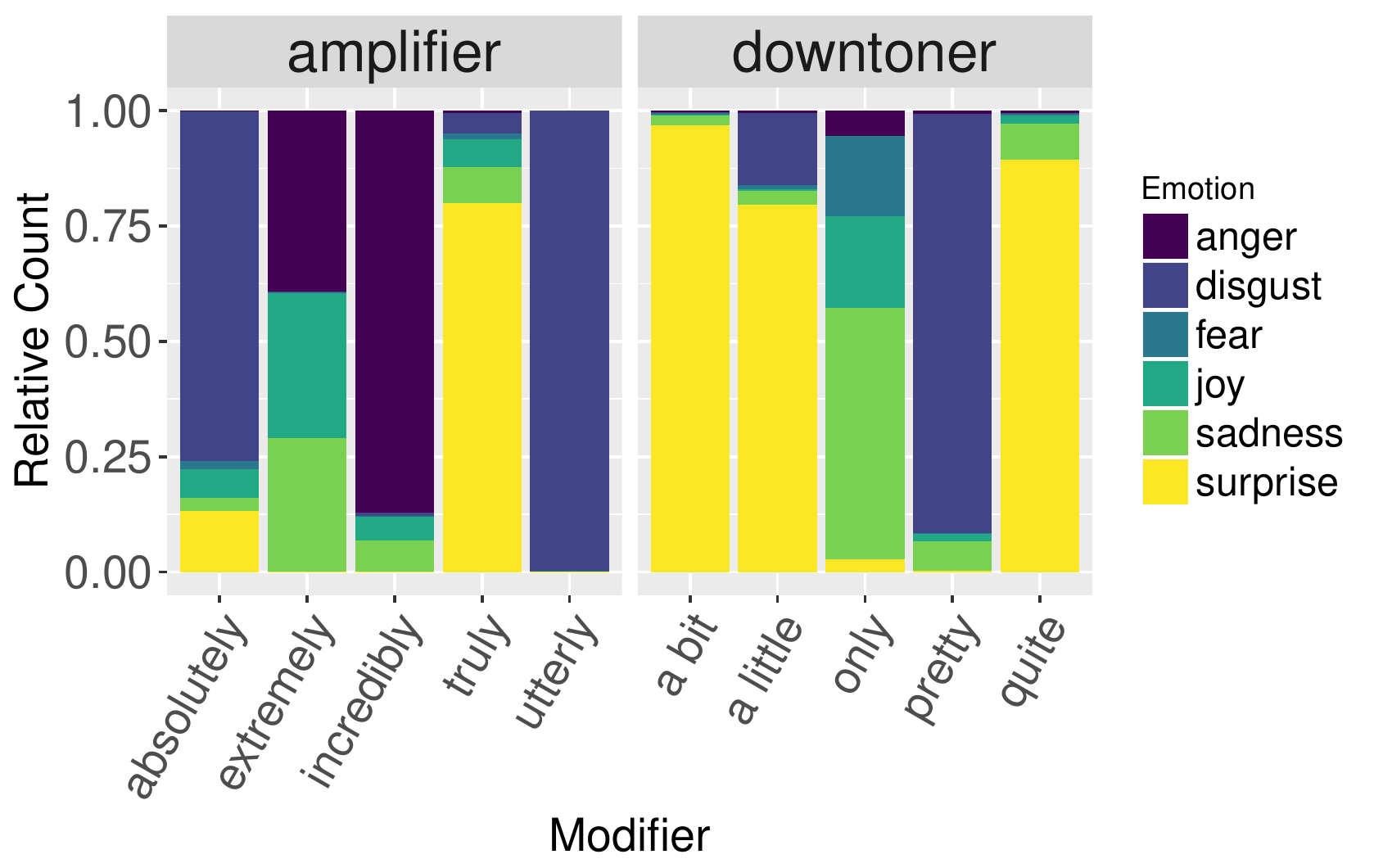}
   \caption{Amplifiers and downtoners that vary the most in use with regards to emotion}
   \label{fig:selection}
\end{figure}

\begin{figure}[t]
   \centering
   \includegraphics[width=0.9\linewidth]{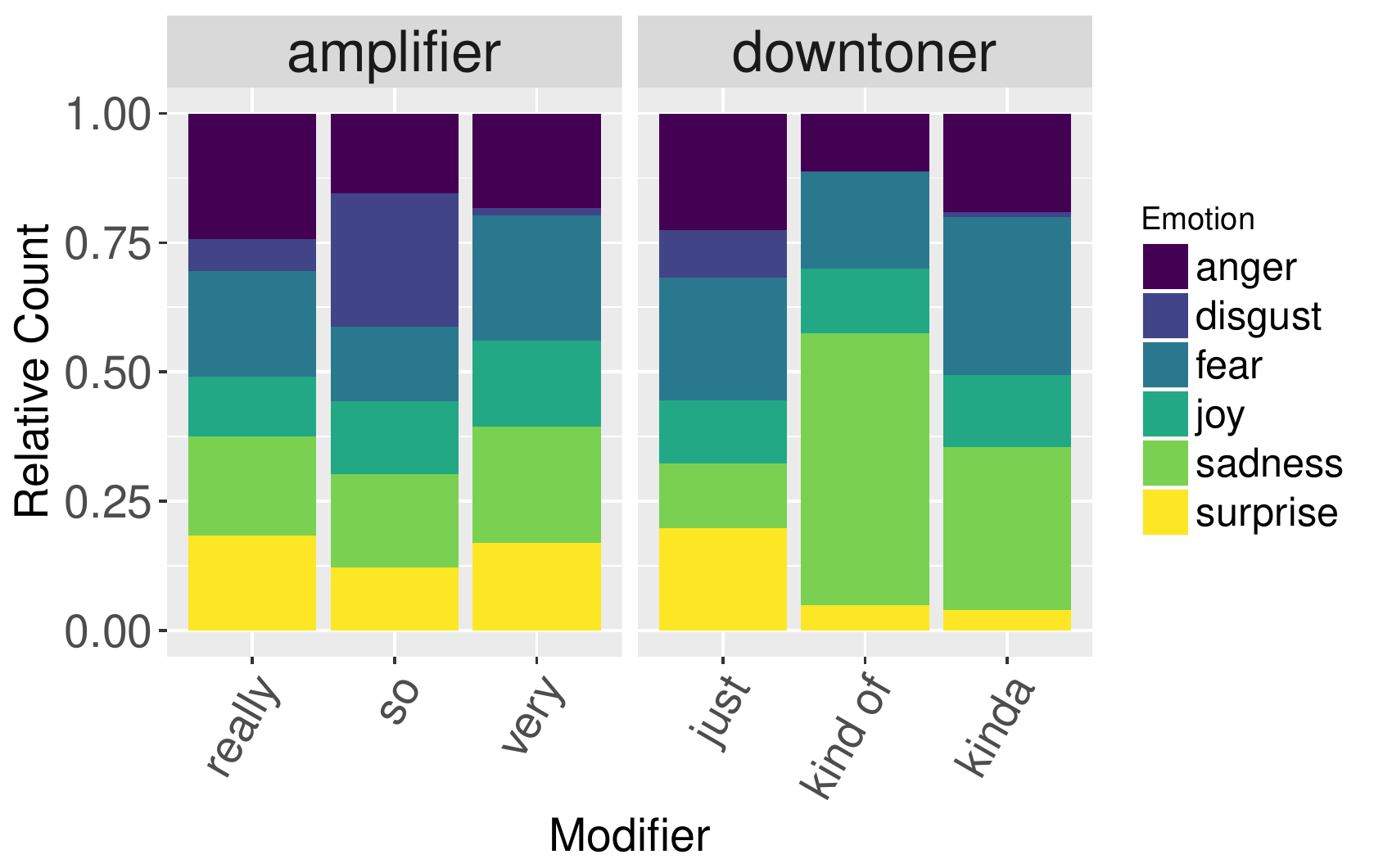}
   \caption{Most frequent three amplifiers and downtoners used across all
emotions and their variation with respect to emotion.} \label{fig:top}
\end{figure}

In Figure~\ref{fig:selection} and Figure~\ref{fig:top} we observe that the use
of modifiers with respect to an emotion vary a lot within the same class of
modifiers among both more frequent and less frequent modifiers. In
Figure~\ref{fig:selection}, we observe that the focus modifier ``only'' scales
downward \textit{surprise} the least, while all the other ``true'' scaling
adverbs are more impactful. \textit{Sadness} is the emotion that is mostly
expressed through the focus adverb ``only''  in this setting. The figure also
(implicitly) shows that certain modifiers prefer certain adjectives, e.g. the
adjectives that express \textit{disgust}, such as ``disgusted'' is mostly
modified by ``absolutely'', ``truly'', ``utterly'', ``pretty'' and not by
``extremely'', ``incredibly'' or ``only''. This distinction show the ``harmony''
between adjectives and degree adverbs \cite{quirk1985contemporary}.

Looking in more depth into the most frequent used amplifiers and downtoners in
Figure~\ref{fig:top} we see that among the top used amplifiers ``so'', ``really'',
``very'' we find that \textit{joy}, \textit{anger}, and \textit{disgust} prefer
``so'' over ``really'' and ``very'', the emotions \textit{fear} and
\textit{surprise} prefer ``very'' over ``so'' and ``really'' and
\textit{sadness} is modified rather equally by the three amplifiers. Between
the downtoners ``kind of'' and ``kinda'' there is a notable difference in use
for \textit{sadness}, \textit{fear} and \textit{anger}, with ``kind of'' being
prefered over ``kinda'' in the context of \textit{sadness}, with the opposite
holding true for \textit{fear}.

\begin{table}[tb]
\centering
\setlength{\tabcolsep}{1pt}
\begin{tabular}{lccc}
\toprule
    & \multicolumn{3}{c}{Spearman's rank correlation} \\
    \cmidrule{2-4}
    Emotion     & \textit{w/ context} & \textit{w/o context} & \textit{between}\\
\midrule
anger      & .84 & .82 & .88 \\
fear       & .84 & .73 & .81 \\
joy        & .90 & .86 & .91 \\
sadness    & .90 & .86 & .88 \\
surprise   & .71 & .71 & .81 \\
disgust    & .86 & .86 & .88 \\
\midrule
average  & .84 &.80 & .86\\
\bottomrule
\end{tabular}
\caption{Split-half reliabilities and Spearman's rank correlation between these settings.}
\label{tab:reliab}
\end{table}

\subsection{Annotation Analysis} \label{sec:resanno}
Table~\ref{tab:queries-examples} shows examples of phrases annotated with
real-valued scores following the annotation procedure described in
Section~\ref{sec:data}. We see that we have scores for each phrase in the
context of each emotion.  For instance, ``kinda surprised'' has the score
$-.37$ for \textit{sadness} and $+.17$. We observe that the negation ``not''
paired with any emotion adjective, excluding \textit{happy} obtains a positive
score for \textit{joy}, and a negative score for every other emotion. The
phrase ``not happy'' obtains a negative score of only $-.52$. In the complete
annotation results we include as negations also the phrase ``not happy at
all'', which in this case gets closer to the lower limit of the potential
scores.

We measure the reliability by randomly dividing the sets of 4 responses to each
question into two halves and comparing the Spearman rank correlation
coefficient between the two sets \cite{kiritchenko2017capturing}. Both with and
without having access to context, the annotators mostly agree regarding their
annotations, as Table~\ref{tab:reliab} shows in the first two columns. Lowest
reliability is achieved for \textit{surprise}, with .71 Spearman's rank
correlation and the highest for \textit{joy} and \textit{sadness} (.9). The
reliability drops most when context is not available for fear (by 11 percentage
points).

 \begin{figure}
   \centering
   \includegraphics[width=0.9\linewidth]{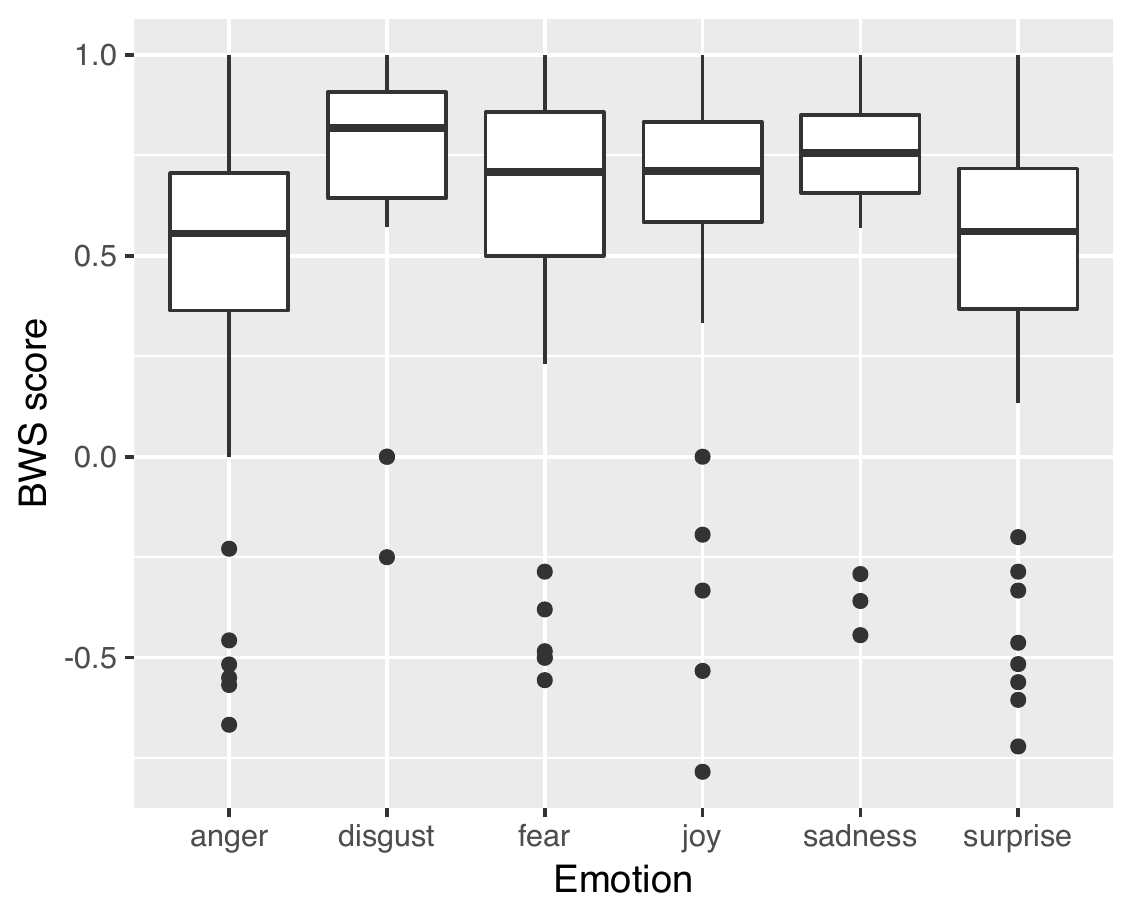}
   \caption{Distribution of the aggregated emotion scores obtained by
     applying the counting procedure BWS%
      }
   \label{fig:boxplot}
 \end{figure}

Figure~\ref{fig:boxplot} shows the distribution of the scores
assigned through the annotation per emotion. We observe that
\textit{disgust} is mostly amplified and rarely negated (only once). The
outliers in each boxplot mostly correspond to negated phrases.

\begin{figure}[t]
  \centering
  \includegraphics[width=1\linewidth]{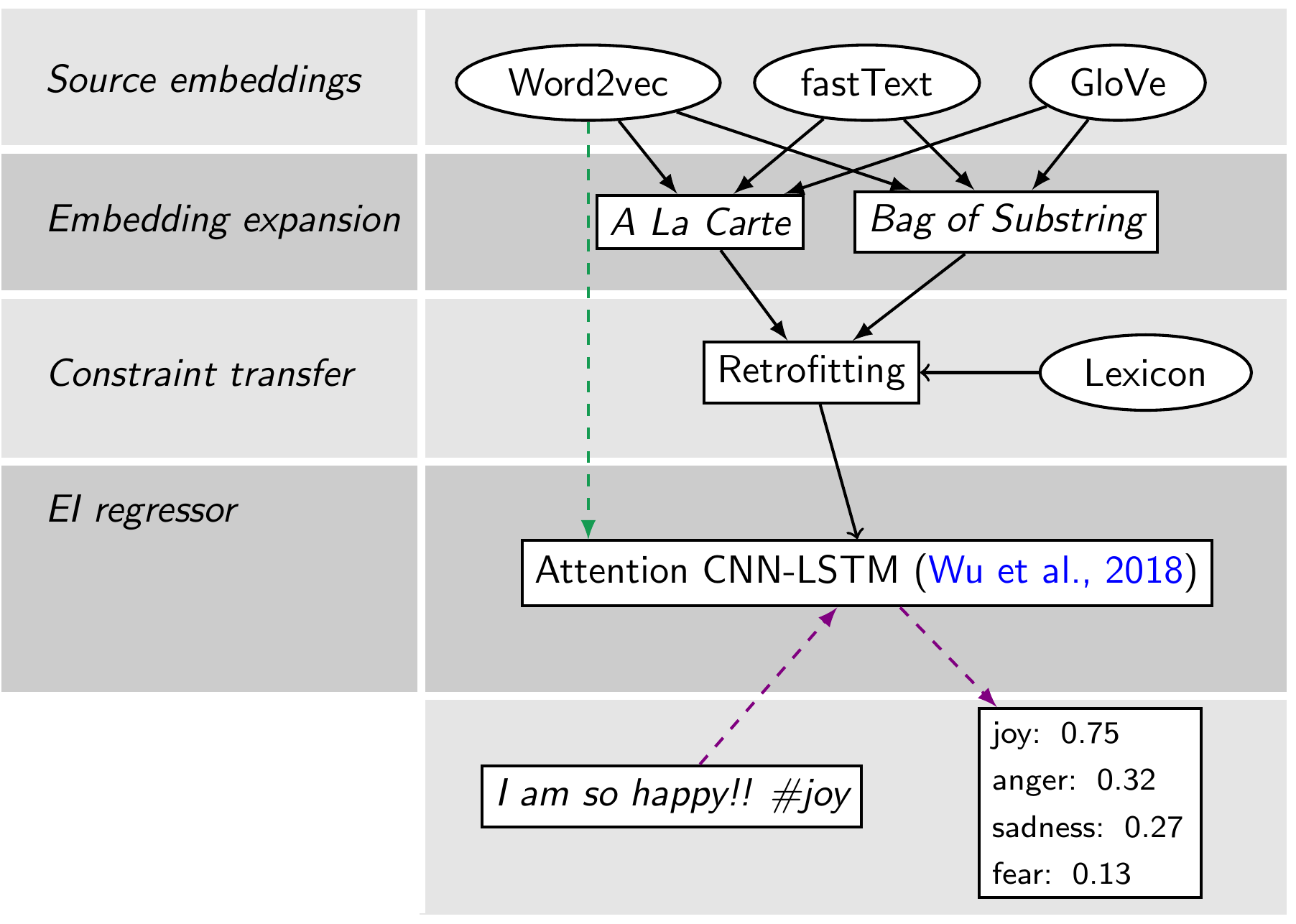}
  \caption{Experimental Setup. The green arrow from Word2vec to
    the regressor unit shows the
    information flow in the baseline. The black solid arrows show the
    different experimental settings. The purple dashed arrows at the
    bottom show the prediction phase.}
  \label{fig:setting}
\end{figure}

\subsection{Embedding Adaptations} \label{sec:adapt}
Figure~\ref{fig:setting} summarizes our experimental setup. We build
on top of pretrained embeddings obtained with \wordtovec{}
\cite{mikolov2013distributed} (300d, negative sampling, Google News
corpus), \fasttext{} \cite{bojanowski2016enriching} (300d, news
corpora), or \glove{} \cite{pennington2014glove} (300d, Common
Crawl). Each embedding is then optionally augmented with phrase and
subword embeddings and fed into a CNN-LSTM model as proposed by
\newcite{wu2018thungn}, trained on the Affect in Tweets Dataset used
at Sem Eval 2018 Task 1 \cite{SemEval2018Task1}. Their system achieved
an average Pearson correlation score of 0.722, and ranked 12/48 in the
emotion intensity regression task.

Table~\ref{tab:downstream} shows Spearman's rank correlation between
the predicted intensity scores and the emotion scores obtained in the
annotation of our Twitter corpus or the EmoInt data
\cite{mohammad2017wassa}.

The \fasttext-based models underperform constantly on our Twitter
dataset. For \glove{} embeddings, \alacarte{} (ALC) and
\bagofsubstrings{} (BoS) lead to a substantial improvement, of 7pp
(see Table~\ref{tab:downstream}, G vs.\ G+ALC) and
8pp (G vs.\ G+BoS) over the baseline of using the pretrained
embeddings unchanged. On \wordtovec{} embeddings BoS and ALC show the
same improvement of 7pp (W2V vs.\ W2V+ALC/BoS). 

While on average, ALC and BoS can only substantially contribute based
on \glove{} and \wordtovec{}, this is not the case for individual
emotions. For \wordtovec, sadness figures to be particularly
challenging, leading to an overall comparably low performance. Most
importantly, we observe that our extensions of the semantic spaces do
not negatively affect the results on the EmoInt dataset.

Unexpectedly, retrofitting does not help in all settings in our
post-processing pipeline except for \fasttext{} embeddings.  We assume
that is a consequence of using a too small lexicon for retrofitting,
and the method would improve the embeddings if sentiment or emotion
lexicons would be used instead. However, this needs further
investigation.

\begin{table}[tb]
    \small
  \centering
  \setlength{\tabcolsep}{1.0mm}
  \newcommand{\glv}{G}
  \newcommand{\alc}{ALC}
  \newcommand{\wordvec}{W2V}
  \newcommand{\retro}{RF}
  \newcommand{\bos}{BoS}
  \newcommand{\ft}{FT}
  \begin{tabular}{lccccccc}
    \toprule
    & \multicolumn{1}{c}{\rt{joy}} & \multicolumn{1}{c}{\rt{sadness}} & \multicolumn{1}{c}{\rt{anger}} & \multicolumn{1}{c}{\rt{fear}} & \multicolumn{1}{c}{\textit{\rt{average}}}\\
    & T \hspace{.2em} EI& T \hspace{.2em} EI& T \hspace{.2em} EI& T \hspace{.2em} EI& T \hspace{.2em} EI\\
    \midrule
    \glv & .20 .60 & .21 .59 & .24 .60 & .27 .61 & .23 .60 \\
    \glv +\alc  & .23 .61 & .31 .63 & .33 .62 &  .35 .62 &  .30 .63\\
    \glv +\bos & .24 .58 & .30 .60 & .34 .59 & .36 .57 & .31.59 \\
    \glv +\alc +\retro  & .19 .60 & .21 .61 & .26 .63 & .28 .61 & .24.61\\
    \glv +\bos +\retro & .19 .62 & .21 .60 & .28 .62 & .25 .61 & .23 .61 \\
    \midrule
    \wordvec   & .16 .60  & .12 .59  & .19 .60 & .23 .63 & .18 .62 \\
    \wordvec +\alc   & .20 .60 & .24 .64 & .28 .65 & .28 .64 & .25 .63\\
    \wordvec +\bos & .20 .61 & .23 .64 & .28 .66 & .29 .60 & .25 .64\\
    \wordvec +\alc +\retro & .21 .60 & .25 .54 & .28 .69 & .28 .64 & .26 .62\\
    \wordvec +\bos +\retro & .16 .60 & .12 .61 & .24 .67 & .20 .60 & .18 .63\\
    \midrule
    \ft & .16 .58  & .14 .53 & .21 .65 & .22 .60 & .18 .61\\
    \ft +\alc   & .16 .59 & .14 .52 & .21 .59 & .23 .62 & .19 .59\\
    \ft +\bos & .16 .60 & .14 .59 & .22 .63 & .23 .61 & .18 .62\\
    \ft +\alc +\retro & .18 .54 & .16 .62  & .22 .64 & .25 .59 & .20 .60\\
    \ft + \bos +\retro & .16 .60 & .14 .57 & .22 .62 & .21 .57 & .18 .63 \\
    \bottomrule
  \end{tabular}
  \caption{Evaluation: Spearman's rank correlation between predicted emotion
    intensity scores and annotated scores on our dataset (T) or the EmoInt
    dataset (EI). We report results only for the 4 emotions annotated in the
    EmoInt data. }
\label{tab:downstream}
\end{table}

\section{Conclusion \& Future Work}
With this paper, we presented the first analysis of the distribution
of degree adverbs and negations on Twitter in the context of
emotions. In addition, we proposed a pipeline with different modules
to expand embeddings particularly for emotion phrases. Our evaluation
shows substantial differences based on the combination of input
embeddings and the postprocessing method. Our pipeline improves the
results obtained while evaluating the downstream task of emotion
intensity prediction on our dataset. Finally, we contribute a novel
emotion phrase lexicon of high precision.

For future work we propose to analyze other baseline approaches,
particularly learning a composition function over pairs of adjectives
with degree adverbs. The modifiers could be considered as functions
over adjectives and would be represented as matrices.

Another further improvement of this work would be to expand this
analysis to verbal and nominal expressions of emotion, which we
hypothesize as also being frequent. In order to obtain meaningful
representations for the phrases we focus on, another natural next step
is expanding the postprocessing pipeline and including a comparison to
other adaptation methods such as counterfitting
\cite{mrkvsic2016counter}. Presumably, this will also generate
additional insights into the aspect that we were only able to show a
limited improvement based on retrofitting.

Given the recent advances in representing contextualized word
embeddings as functions computing dynamically the embeddings for words
given their context, we hypothesize and intend to further verify that
these embeddings would be a better choice for input to systems that
predict intensity scores.  It would be interesting to compare models
such as word embeddings from language models (Elmo) \cite{elmo},
bidirectional encoder representations from transformers (BERT)
\cite{bert}, and generative pre-training OpenAI (GPT)
\cite{radford2019} to the ones we already discussed, since the
contextualized embeddings assign a different vector for a word in each
given context. These approaches presumably produce a different vector
for ``happy'' in the context of ``not'' than in the content of
``very'' or ``completely''.

Lastly, we plan to also adjust the lexica created such that it covers
more domains, sources, and languages.

\section*{Acknowledgements}
This research has been funded by the German Research Council (DFG),
projects SEAT (Structured Multi-Domain Emotion Analysis from Text, KL
2869/1-1). We thank Jeremy Barnes, Evgeny Kim, Sean Papay, Sebastian
Pad\'o and Enrica Troiano for fruitful discussions and the reviewers
for the helpful comments.

\end{document}